\documentclass[runningheads,a4paper,english]{llncs}[2022/01/12]
\pdfoutput=1
\authorrunning{Victor Christen et al.}
\usepackage{graphicx}
\usepackage{balance}  % for  \balance command ON LAST PAGE  (only there!)
\usepackage[hidelinks=true, colorlinks=true, citecolor=blue, allcolors=blue]{hyperref}
\usepackage{cleveref}
\usepackage{booktabs}
\usepackage[super]{nth}
\usepackage{comment}
\usepackage[numbers,sort]{natbib}
\usepackage{algorithm2e}
\usepackage{multirow}
\usepackage{multicol}
\usepackage{subcaption}
\usepackage{enumitem}
\usepackage{color}
\usepackage{xurl}
\usepackage{orcidlink} % for orcid image
\title{Graph-based Active Learning for Entity Cluster Repair}

\author{Victor Christen\hspace{0.5mm}\orcidlink{0000-0001-7175-7359} \and Daniel Obraczka\hspace{0.5mm}\orcidlink{0000-0002-0366-9872}
\and Marvin Hofer\hspace{0.5mm}\orcidlink{0000-0003-4667-5743}
\and Martin Franke\hspace{0.5mm} 
\and Erhard Rahm\hspace{0.5mm}\orcidlink{0000-0002-2665-1114}}
\institute{Leipzig University \& ScaDS.AI Dresden/Leipzig, Germany\\
            \email{\{christen, obraczka, hofer,franke, rahm\}@informatik.uni-leipzig.de}}
\begin{document}

\maketitle
\begin{abstract}
Cluster repair methods aim to determine errors in clusters and modify them so that each cluster consists of records representing the same entity. Current cluster repair methodologies primarily assume duplicate-free data sources, where each record from one source corresponds to a unique record from another. However, real-world data often deviates from this assumption due to quality issues. Recent approaches apply clustering methods in combination with link categorization methods so they can be applied to data sources with duplicates. Nevertheless, the results do not show a clear picture since the quality highly varies depending on the configuration and dataset. In this study, we introduce a novel approach for cluster repair that utilizes graph metrics derived from the underlying similarity graphs. These metrics are pivotal in constructing a classification model to distinguish between correct and incorrect edges. To address the challenge of limited training data, we integrate an active learning mechanism tailored to cluster-specific attributes. 
The evaluation shows that the method outperforms existing cluster repair methods without distinguishing between duplicate-free or dirty data sources. Notably, our modified active learning strategy exhibits enhanced performance when dealing with datasets containing duplicates, showcasing its effectiveness in such scenarios.
\end{abstract}
\section{Introduction}\label{sec:intro}

The upcoming trend of systems using generative large language models tremendously simplifies knowledge acquisition, being relevant for many industries such as economic companies, medicine, and research. However, the performance of these language models highly depends on the availability of large corpora. Small and specific domains lack the availability of such large corpora. Recent work proposes the usage of knowledge graphs in combination with large language models to overcome these issues~\cite{Pan23KGLLM, Yang23GPTKG}. 

Therefore, constructing knowledge graphs is an essential cornerstone to improve such systems based on these models~\cite{Hofer2023ConstructionOK}. Moreover, knowledge graphs enable an integrated view of data from different heterogeneous data sources, enhancing comprehensive analysis. 

Due to the huge amount of available data sources, data integration is required to structure data into knowledge graphs. 
In recent decades, many record linkage methods have been proposed to identify records from different data sources representing the same entity. In the case of multiple data sources, the result of a record linkage method is a set of clusters where each cluster consists of records representing the same entity. However, due to the heterogeneity and various data quality problems, the resulting clusters might not be correct regarding the matched records. Therefore, in current research, clustering repair methods were investigated to improve the cluster quality by modifying the clusters. The majority of proposed methods~\cite{saaedi2018famer,nentwig2017distributed,ngomo2014collibri} assume that the data sources are duplicate-free and intensively utilize that condition to remove links so that the resulting clusters satisfy the assumption.

However, the assumption that there are no duplicates is unrealistic for a large number of data sources being available in the LOD cloud. Consequently, methods that intensively use this assumption achieve poor results. Recent works apply modified clustering methods such as hierarchical clustering or affinity propagation~\cite{SaeediDR2021, LermSR2021} that can be used to repair clusters from dirty data sources. Nevertheless, the results highly depend on the configuration with respect to a certain linkage problem. 

To overcome such issues, we suggest a novel repair approach that utilizes graph metrics as indicators for correct and incorrect links to repair clusters based on generated similarity graphs. Specifically, we make the following contributions:
\begin{itemize}
    \item We propose a cluster repair method based on a classification model using graph metric-based features. In addition to the similarities, the used features cover network information within a cluster. The repair step utilizes the model to iteratively add or delete records from clusters depending on the prediction and the relationship to the existing records of a cluster. 
    \item Due to the lack of training data, an active learning method is integrated into our method. To generate representative training data regarding the different clusters with their specific characteristic, we extend an existing active learning method by considering cluster-specific features in the selection phase. 
    \item We intensively evaluate our approach regarding the used labeling budget and selection strategies on two real-world datasets. We compare the results with existing cluster repair methods focusing on duplicate-free and dirty data sources. In the end, we also verify the robustness according to noisy similarity graphs by changing the similarities of edges randomly.
\end{itemize}

The remainder of this paper is structured as follows. In~\Cref{sec:problem_def}, we define the problem of repairing clusters of records and discuss related work (\cref{sec:related_work}).   In~\Cref{sec:method}, we present our novel approach for repairing clusters utilizing graph metrics with active learning.  In~\Cref{sec:eval}, we evaluate our approach on different datasets to validate its practical applicability. Finally, we conclude our work in~\Cref{sec:conclusion}.
\begin{figure}[t]
     \centering
     \includegraphics[width=\textwidth]{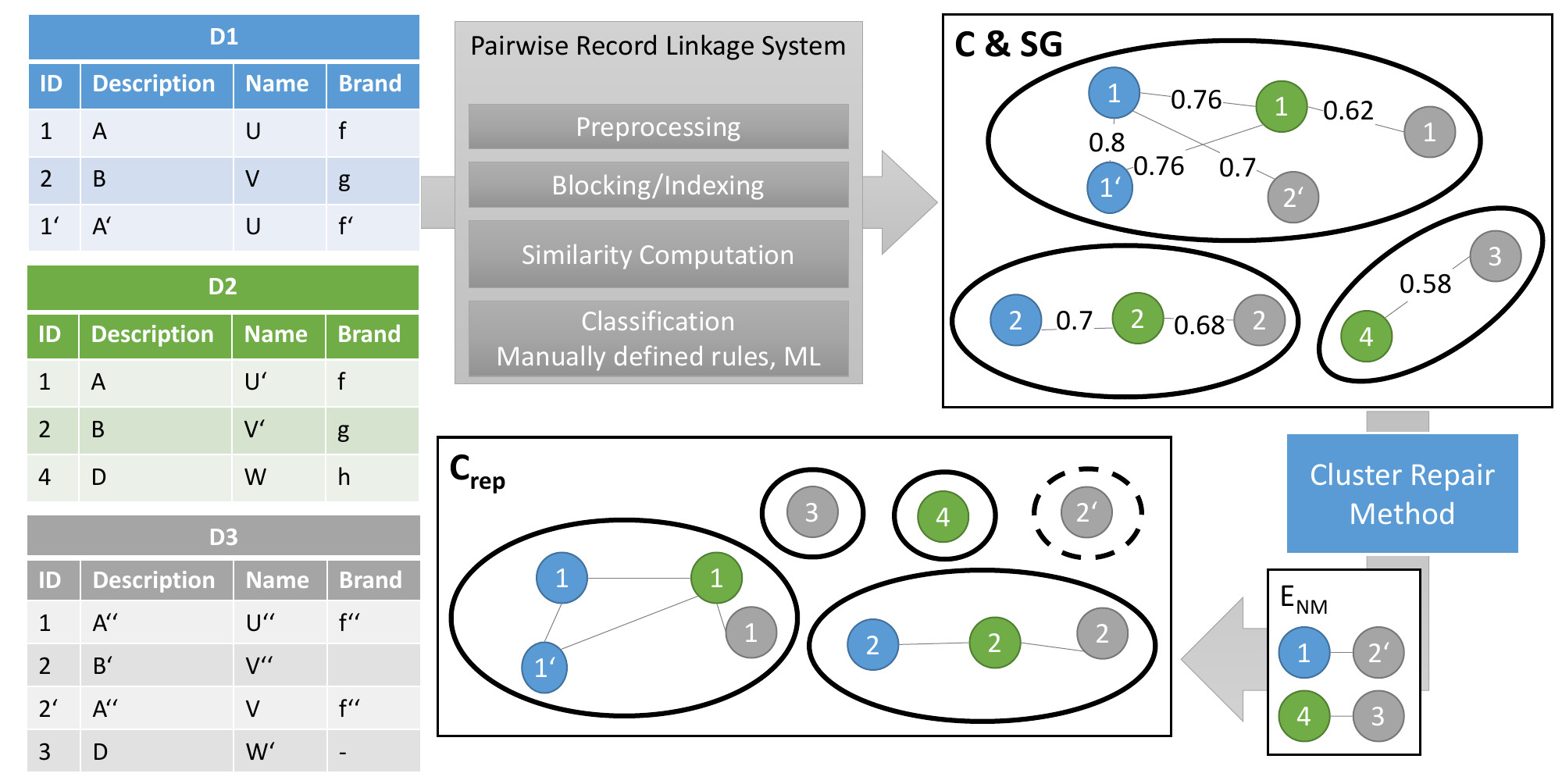}
     \caption{Outline of the complete entity resolution process including the repair method identifying incorrect edges $\mathbf{E_{NM}}$ to construct repaired clusters $\mathbf{C_{rep}}$.}
     \label{fig:problem_def}
\end{figure}
\section{Problem Definition}\label{sec:problem_def}
Let $\mathcal{D}$ be a set of data sources $D_1$, $D_2$, ...$D_n$ and $SG$ a similarity graph generated by an arbitrary entity resolution tool, exemplary depicted in \Cref{fig:problem_def}, or derived from a collection of \texttt{sameAs} links. The entity resolution tool generates for each data source pair a set of \texttt{sameAs} links between records by executing the typical steps: preprocessing, blocking, similarity computation, and classification~\cite{Chr12}. Note, that the similarity graph is not complete because of the classification step so that not each record pair represents an edge. Due to the transitivity of \texttt{sameAs} links, the records of a connected component represent a cluster $c$ and thus the same entity. 

A repair method aims to determine incorrect clusters consisting of wrongly assigned records according to the initial set of clusters $\mathbf{C}$ and update them to a repaired set of clusters $\mathbf{C_{rep}}$. To identify incorrect records, the approach can utilize the computed similarity graph $\mathbf{SG}$ with its similarities to remove incorrect links $\mathbf{E_{nm}}$ resulting in a partition of the connected component. In general, a record linkage approach computes multiple similarities regarding different attributes. In this work, we assume that the similarities are aggregated by an aggregation function. Moreover, a similarity can also be a probability of a classification model of how likely the match is. Due to the high complexity of the pairwise comparison, we do not consider missing merged clusters such as the cluster of r2' with the cluster consisting of the other records r2.

%TODO weil du typical schreibst. Sollte man nicht noch erwähnen das blocking (oder) ein threshold der grund ist das der similarity graph nicht similarities von jedem zu jedem knoten aufweist?
\section{Related Work} \label{sec:related_work}

Constructing knowledge graphs essentially depends on entity resolution, a field extensively studied for decades~\cite{Chr12, NentwigHNR17survey}. The primary objective lies in identifying records that represent the same entity. Most of the approaches~\cite{Ngonga21limes, Doan20magellan, Mudgal2018deepLearning} as well as recent research utilizing large language models~\cite{Bing21Bert, Peters2023GPT} treat entity resolution as a classification task, categorizing record pairs as matches or non-matches. Due to the quality issues and the transitivity of \texttt{sameAs} links, a pairwise view is not sufficient to generate qualitative record clusters. The error-prone clusters can lead to a wrong knowledge graph construction and lead to relationships failing to represent the real world.

As a solution, methods for cluster repair become relevant to enhance the quality of derived clusters, building upon the determined \texttt{sameAs} links from any entity resolution method.

\textbf{Multi-source entity resolution:} Due to the increasing heterogeneity considering data sources and the potentially high number of data sources, multi-source entity resolution methods address these challenges. Chen et al.~\cite{shen08Soccer} define a matching plan that specifies which matcher is applied to which source and which sources are grouped. In addition to the matching task, a repair step is required that identifies conflicts introduced by the transitive closure and resolves them. CLIP~\cite{saaedi2018famer} categorizes the computed links into \emph{strong}, \emph{normal}, and \emph{weak} links based on the similarity graph structure and the data sources. The different link categories distinguish if two records from two data sources are connected by an edge that maximizes the similarity for both records or not. Using the link categories and the assumption of duplicate-free data sources, the method iteratively removes edges from clusters until they are source-consistent. Lerm et al.~\cite{LermSR2021} extended affinity propagation clustering applicable for clean as well as dirty data sources. Experiments showed that the approach leads to small clusters with high precision but at the cost of relatively low recall. To overcome the low recall, Saeedi et al.(2021)~\cite{SaeediDR2021} proposed an agglomerative hierarchical clustering-based method using the basic strategies: \emph{single-}, \emph{complete-}, and \emph{average-}linkage. To guarantee source consistency regarding clean data sources, they adapted the general method by adding certain constraints for merging clusters.

\textbf{Active Learning:}
The goal of active learning methods is the interactive and iterative training data generation by determining informative unlabeled samples. The selected data will then be classified by an oracle and added to the existing training data. In each iteration, the training data is utilized to determine new informative samples. The algorithm terminates if a specific stop criterion is achieved such as the number of manually classified samples so-called labeling budget or the performance of the current classification model is sufficient. Due to the lack of available training data, various methods~\cite{Ara10, Bel12, Ngo12, christen19infoal, Moz14} have been proposed for entity resolution. The works of~\cite{Ara10, Bel12} aim to maximize the recall regarding a specified precision threshold. EAGLE~\cite{Ngo12} focuses on RDF link discovery and aims at generating link specifications representing a complex similarity function. Mozafari et al.~\cite{Moz14} proposed two such approaches, named \emph{Uncertainty} and \emph{MinExpError}, being applicable for applications beyond entity resolution. The main idea of these approaches is to use non-parametric bootstrapping to estimate the uncertainty of classifiers. Christen et al.~\cite{christen19infoal} defined an informative measure for selecting new samples based on the location of samples in the similarity vector space instead of generating a model in each iteration. However, the proposed methods only focus on linkage problems between two data sources. Recent work~\cite{Primpeli21graphAl} also addresses mutli-source entity resolution problems introducing new challenges such as increasing heterogeneity and search space regarding informative samples. 

The mentioned active learning methods do not address the challenges of heterogeneity introduced by various data sources. Recently, Primpeli et al.~\cite{Primpeli21graphAl} proposed a novel approach utilizing graph characteristics based on similarity graphs to determine informative links. The advantage of this method is that it likely selects links leading to a wrong-connected component. The results showed that the approach outperforms existing active learning methods based on a pairwise view using a small labeling budget.

\begin{figure}[tb]
     \centering
     \includegraphics[width=\textwidth]{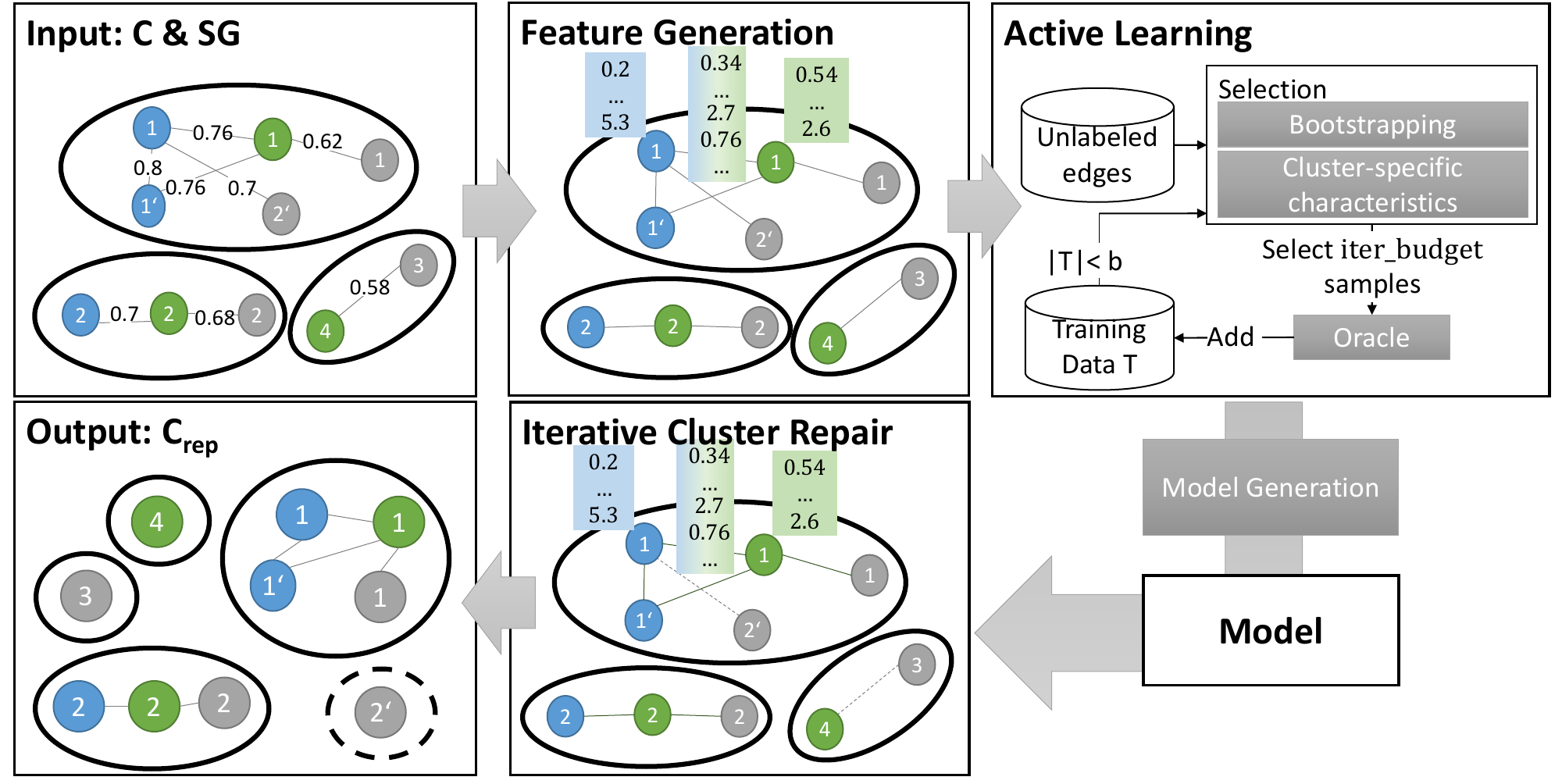}
     \caption{Overview of the graph-based cluster repair method.}
     \label{fig:overview_method}
\end{figure}
\section{Graph-based Cluster Repair}\label{sec:method}
In the following, we will present our graph-based cluster repair method depicted in \Cref{fig:overview_method}.
The approach aims to generate a set of clusters $\mathbf{C_{rep}}$ consisting of records representing the same entity. As input the method has an initial set of clusters $\mathbf{C}$ derived from a similarity graph $\mathbf{SG}$ generated by an arbitrary record linkage framework. Each connected component forms a cluster $c$ with records that should represent the same entity. The similarity graph $\mathbf{SG}$ is a weighted undirected graph consisting of nodes representing records and edges indicating the similarity between two records. The similarity can be a probability determined by a machine learning model or an aggregated similarity regarding different attribute similarities. Similarly to CLIP~\cite{saaedi2018famer}, we also remove edges if both adjacent records are related to edges with a higher similarity.

Due to data quality issues and linkage errors, the resulting clusters derived from a similarity graph might consist of records that are not the same as the other records of the cluster. In contrast to pairwise record linkage, edges can be characterized by the adjacency nodes and the indirectly connected nodes. For representing the relationship of connected records, we compute various graph metrics in the feature generation step such as PageRank, and several centrality measures to construct node and edge features. 

To determine the correctness of an edge, we classify each edge based on a trained model $\mathbf{M}$ using the derived edge features. The training process requires classified edges as matches and non-matches. Due to the sparsity of available training data, we adapt an existing active learning~\cite{Moz14} to select informative edges as training data $\mathbf{T}$. In each iteration, we select $iter\_budget \geq 0$ unlabeled samples for labeling and add the labeled edge feature vectors to the existing training dataset $\mathbf{T}$. The iteration terminates if the specified labeling budget $b$ is exhausted. We extend the approach by considering cluster-specific characteristics such as the number of nodes. The intention is that the selected training data should represent the different available clusters. Therefore, the cluster-specific properties of the selected edges should be similar to the characteristics of all clusters. In the last step, we iteratively repair the initial clusters $\mathbf{C}$ using the model $\mathbf{M}$ so that each group consists of records where the edges are classified as \emph{match} with a high support. 

\subsection{Feature Generation}
A classification model classifies the edges in a cluster using graph metric-based feature vectors to determine if the edges are correct. The link categories proposed in \cite{saaedi2018famer} and the similarities characterize edges directly. In addition to local edge properties, it is also represented by network information regarding the induced similarity graph of a cluster. 
The set of features forms a vector $\overrightarrow{e}$ for each edge used in the classification step.
Due to the same structure of different clusters, various edges can be represented by the same edge vector $\overrightarrow{e}$. Therefore, we only consider unique edge vectors in the training phase for generating the classification model $\mathbf{M}$. We select several metrics~\cite{Newman2010NetworksAI, Harris08bridges} for feature computation to characterize an edge by the graph's element type and the information type.  \Cref{tab:features} shows an overview of the features.
\setlength{\tabcolsep}{10pt}
\begin{table}[]
    \centering
    \begin{tabular}{lcc}
        \toprule
         name & element type & information\\
         \midrule
         PageRank             & node &network\\
         Closeness Centrality  & node &network\\
         Betweenness Centrality& node &network\\
         Cluster Coefficient   & node &network\\
         Similarity            & edge &local\\
         Link category         &  edge &network\\ 
         Bridge                & edge &network\\
         Betweenness Centrality& edge &network\\
         Complete ratio ($|E|/\frac{|V|\cdot(|V|-1)}{2}$)   & graph&network\\
         \bottomrule
    \end{tabular}
    \caption{Overview of computed features to characterize an edge categorized by the element type of a graph and the type of information.} \label{tab:features}
\end{table}

\subsection{Cluster Characteristic Aware Active Learning}
Due to the high number of clusters and the lack of evaluated clusters, we need to generate training data efficiently. In our case, the training data consists of edge feature vectors labeled as correct or not. Active learning techniques enable an efficient and interactive selection of informative samples. The current research provides various methods for record linkage. However, most of them focus on the training data selection to train a record linkage classification model based on attribute value similarities. In our case, we also want to consider cluster-specific characteristics to select edge feature vectors as training data. Therefore, we extend the active learning method of Mozafari et al.~\cite{Moz14} by considering the number of nodes regarding the cluster of the selected edge. The main idea is to select iteratively a certain number $iter\_budget$ of unlabeled edge feature vectors being informative to extend the current training data $\mathbf{T}$. The iterative process terminates if a stop criterion such as a total labeling budget $b$ is reached. To determine informative vectors Mozafari et al. use a bootstrapping technique. Therefore, the method generates $k$ classifiers based on the current training dataset $\mathbf{T}$ by sampling with repetition. 
The determined models $m_1,..,m_k$  classify the unlabeled edge feature vectors where the predictions are utilized to compute the uncertainty $unc(\overrightarrow{e})$ of an edge $e$ shown in \ref{eq:base_unc}. 
\begin{equation} \label{eq:base_unc}
    unc(\overrightarrow{e}) = \frac{\sum_{i=1}^k m_i(\overrightarrow{e})}{k} \cdot (1-\frac{\sum_{i=1}^k m_i(\overrightarrow{e})}{k})
\end{equation}
The term $m_i(\overrightarrow{e})$ results in 0 or 1 if the edge $e$ represents a non-match, respectively, a match. 

We extend the uncertainty criterion by using the number of nodes regarding the graphs of each edge vector $\overrightarrow{e}$ as a cluster-specific characteristic. Note, that multiple clusters can contain edges with the same edge vector due to the same structure and similarity. The goal is to avoid an over and under-representation of clusters with a certain size concerning all clusters. Consequently, an edge vector can be assigned to more than one cluster-specific characteristic. For selecting an edge vector $\overrightarrow{e}$, we determine a cluster-specific weight $w_c(\overrightarrow{e})$ based on the cluster size distribution from all available clusters $d_{C}$ and the current training data distribution $d_T$. The distribution of $d_{C}$ and $d_{T}$ are represented as vectors with a dimension equal to the maximum number of nodes considering all clusters. An entry of $d_{C}$ and $d_{T}$ consists of the ratio between the frequency of clusters with a certain number of nodes and the total number of clusters in $\mathbf{C}$ resp. $\mathbf{T}$. The weight according to a certain cluster size is computed by the difference $\overrightarrow{w}=d_{C} - d_{T}$. Due to the n:m relationship between edge feature vectors and clusters, we determine the average weight according to an edge feature vector $\overrightarrow{e}$ using the specific weights $w_l$ of $w$ where the $l$-th entry corresponds to the cluster size of a cluster $c$ where $e \in c$ holds. 

In addition to the cluster-specific characteristic, we extend the selection strategy by using the average cosine distance between the unlabeled edge feature vector $\overrightarrow{e}$ and the already selected edge vectors $e_T \in T$. The intention is to select dissimilar vectors compared to the current training dataset. Summarizing, the different measures $unc(\overrightarrow{e})$, $w(\overrightarrow{e})$ and $avg\_cos(\overrightarrow{e})$ are averaged to a final informativeness score being used to order the edge feature vectors. Using the order, we select $iter\_budget$ edge feature vectors in each iteration until the training data size achieves the labeling budget. 

\begin{figure}[th]
    \begin{subfigure}[t]{0.15\textwidth}
     \includegraphics[width=\textwidth]
     {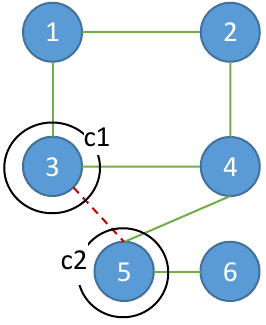}
     \caption{start}
    \end{subfigure}
    \begin{subfigure}[t]{0.25\textwidth}
     \includegraphics[width=\textwidth]
     {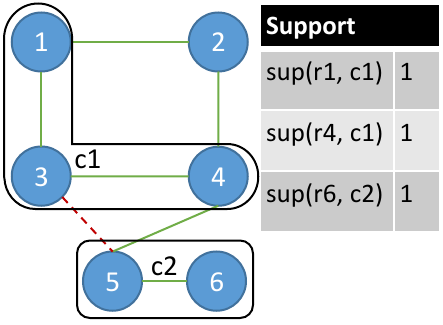}
     \caption{iteration=1}
    \end{subfigure}
    \begin{subfigure}[t]{0.25\textwidth}
    \includegraphics[width=\textwidth]{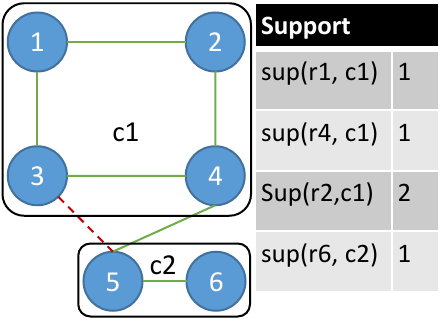}
     \caption{iteration=2}
    \end{subfigure}
    \centering
    \begin{subfigure}[t]{0.25\textwidth}
    \includegraphics[width=\textwidth]{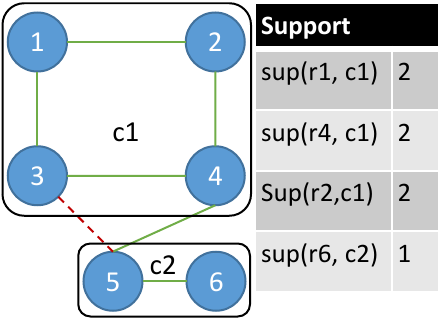}
     \caption{iteration=3}
    \end{subfigure}
    \caption{Example of the iterative cluster repair procedure showing 6 records of an initial cluster. The dashed red line is an edge being classified as non-match.}
    \label{fig:it_example}
\end{figure}

\subsection{Iterative Cluster Repair}

The repair method resolves each cluster utilizing the generated classification model $\mathbf{M}$. At first, we determine for each cluster $c$ the set of edges $E_{NM}$ and $E_{M}$ classified as non-matches resp. as matches using the related edge feature vectors. Each record $u$ and $v$ of an edge $(u, v)\in E_{NM}$ represent a repaired cluster $c_u$ and $c_v$ of $C_{rep}$ because $u$ and $v$ are classified as different entities. We iteratively merge the remaining records into the existing clusters until they are stable. The process utilizes a support value $sup(u,c)$ indicating the strength of the assignment of record $u$ and the cluster $c$. The support is defined as the difference concerning the number of predicted matches and non-matches between the record $u$ and records $v \in c$ being already added to cluster $c$.
The merging process checks each cluster $c \in C_{rep}$ by considering the adjacent nodes $u$ of a record $v \in c$. If the adjacent record $u$ is not connected via an edge predicted as non-match and is not merged into another cluster $c'$, we add the adjacent record $u$ to $c$. We also maintain the support $sup(u,c)$ of $u$ being an element of $c$.  If the record $u$ is already assigned to another cluster c', we compare the support values $sup(u, c)$ and $sup(u, c')$. We change the cluster assignment, if the support $sup(u, c)$ of the considered cluster $c$ is higher than the support $sup(u,c')$ of the previous cluster $c'$. The assignment could differ depending on the order in which the adjacent records are processed. Therefore, we repeat this procedure for the new clusters until they do not change.

\textbf{Example:} \Cref{fig:it_example} shows an example of the iterative cluster repair step. In this example, we have a cluster with 6 records. The classification model classifies the edge $(r3,r5)$ as non-match. Consequently, the records $r3$ and $r5$ represents two clusters $c1$ and $c2$. In the first iteration, the approach considers the adjacent records $r1$ and $r4$ according to $r3$ as well as $r6$ connected to $r5$. Note, that $r4$ will also be considered for $r5$.
The records $r1$ and $r4$ are assigned to cluster $c1$ with a support of 1. Considering the adjacent nodes of $r5$, only $r6$ is assigned to the cluster $c2$. The assignment of $r4$ does not change because the support $sup(r4, c1)=1$ is equal to $sup(r4,c2)=1$. In the second iteration, the assignment of the records $r1$ and $r4$ are the same. The record $r2$ is added to the cluster $c1$ with $sup(r2,c1)=2$. In the next iteration, the assignment does not change so the approach terminates.   
\section{Evaluation}\label{sec:eval}

We evaluate our graph metric-based cluster repair method considering two datasets, namely MusicBrainz and Dexter. Initially, we analyze the impact of the cluster-specific characteristics for selecting new training data in the active learning step. Moreover, we compare the achieved results with existing repair methods utilizing the similarities and link categories discussed in \Cref{sec:related_work}. Due to the challenges of linking, we also evaluate how robust our method is regarding noisy similarity graphs. To measure the effectiveness, we compute the F1-score, defined as the harmonic mean of precision and recall. In pre-experiments, we determined the number of labels selected for each iteration $iter\_budget=20$ and the number of models $k=100$ for determining the uncertainty. We repeated the experiment three times.

In the following, we describe the datasets in \cref{subsec:data}. After that, we compare the impact of our modification of the active learning procedure in~\cref{subsec:al_mod}. We compare our method with existing approaches in~\cref{subsec:comp} and finally, we verify the robustness in~\cref{subsec:noise}.

\begin{table*}[t]
    \caption{Dataset overview and linking configuration of MusicBrainz and Dexter.}
    \label{tab:music_config}
    \centering
    \scriptsize
    \begin{tabular}{c c c c c }
          \toprule
          \multirow{2}{*}{\textbf{Dataset}}&\multirow{2}{*}{MusicBrainz} & \multicolumn{3}{c}{Dexter} \\
          &   &  C0 &  C50 &  C100 \\
          \midrule 
         \textbf{\#Records} &20,000 & \multicolumn{3}{c}{21,023}\\
         \midrule
         \textbf{\#Matches} & 16,250& 368,546 & 38084 & 16014\\ 
         \midrule
         \multirow{2}{*}{\textbf{Attributes}} & Artist, title, album, year,  &\multicolumn{3}{c}{\multirow{2}{*}{Heterog. key-value pairs}} \\
          & length, language, number&\\
          \midrule
         \textbf{Blocking Key} & preLen1(album) & \multicolumn{3}{c}{mfr. name, model number} \\ 
         \midrule
         \multirow{3}{*}{\textbf{Similarity}} & \multirow{5}{*}{Trigram(title)} & \multicolumn{3}{c}{Trigram(model names,}\\ 
         & &  \multicolumn{3}{c}{product code, sensor type),}\\     
         \textbf{Function}& &  \multicolumn{3}{c}{Euclid(opt./digital zoom,}  \\
         & &  \multicolumn{3}{c}{camera dim.,price,}\\
         &&\multicolumn{3}{c}{weight, resolution)}\\
         \bottomrule
     \end{tabular}
    %Trigram(model names, product code, sensor type),Euclid(opt./digital zoom, camera dim., price, weight, resolution) 
\end{table*}

\subsection{Datasets}\label{subsec:data}

We use datasets from two domains: records about music albums (MusicBrainz) and consumer products of type camera (Dexter). Both datasets are multi-source datasets and heterogeneous regarding their error characteristics. In contrast to the MusicBrainz dataset being duplicate-free, the camera dataset is dirty and contains intra-source duplicates.

\subsubsection{MusicBrainz}

The MusicBrainz dataset is a synthetically generated dataset from the MusicBrainz (\url{https://musicbrainz.org/}) database. The dataset is corrupted by~\cite{Hildebrandt2020}, consisting of five sources with duplicates for $50\%$ of the original records. Each data source is duplicate-free, but the records are heterogeneous regarding the characteristics of attribute values, such as the number of missing values, length of values, and ratio of errors. The similarity graphs we used in our evaluation have been utilized in several previous studies~\cite{saaedi2018famer, SaeediER2020, LermSR2021, SaeediDR2021}. The linkage configuration is shown in~\Cref{tab:music_config}. 

\subsubsection{Dexter}

This dataset is derived from the camera dataset of the ACM SIGMOD 2020 Programming Contest (\url{http://www.inf.uniroma3.it/db/sigmod2020contest/index.html}). The dataset consists of 23 sources with $\approx$21,000 records and intra-source duplicates. Each data source consists of source-specific attributes. We used the same linkage configuration as in previous studies~\cite{SaeediDR2021,SaeediER2020} (see \Cref{tab:music_config}).
In addition to the original dataset, Lerm et al.~\cite{LermSR2021} also derive various datasets of different duplicate ratios. Therefore, they deduplicated a specified set of data sources and selected a certain ratio of records to construct clean data sources. The remaining data sources were used to generate the dirty data sources. We consider the datasets $C0, C50$, and $C100$ in our evaluation. For instance, the dataset $C50$ consists of $\approx 50\%$ records from deduplicated sources according to the total number of records of the constructed dataset.

% \begin{figure}[!t]
%     \centering
%     {\includegraphics[width=0.99\textwidth]{figures/MusicBrainz.pdf}}
%     {\includegraphics[width=0.99\textwidth]{figures/C0.pdf}}
%     {\includegraphics[width=0.99\textwidth]{figures/C50.pdf}}
%     {\includegraphics[width=0.99\textwidth]{figures/C100.pdf}}
%     %\vspace{-0.115cm}
%     \caption{Results on Music Brainz and Dexter(C0, C50, C100) datasets with different duplicate ratios considering the basic selection strategy (bootstrap) and the cluster-specific selection (bootstrap ext) in the active learning step.}
%     \label{fig:al_results}
% \end{figure}

\begin{figure}[htpb]
    \centering
    \includegraphics[width=\textwidth]{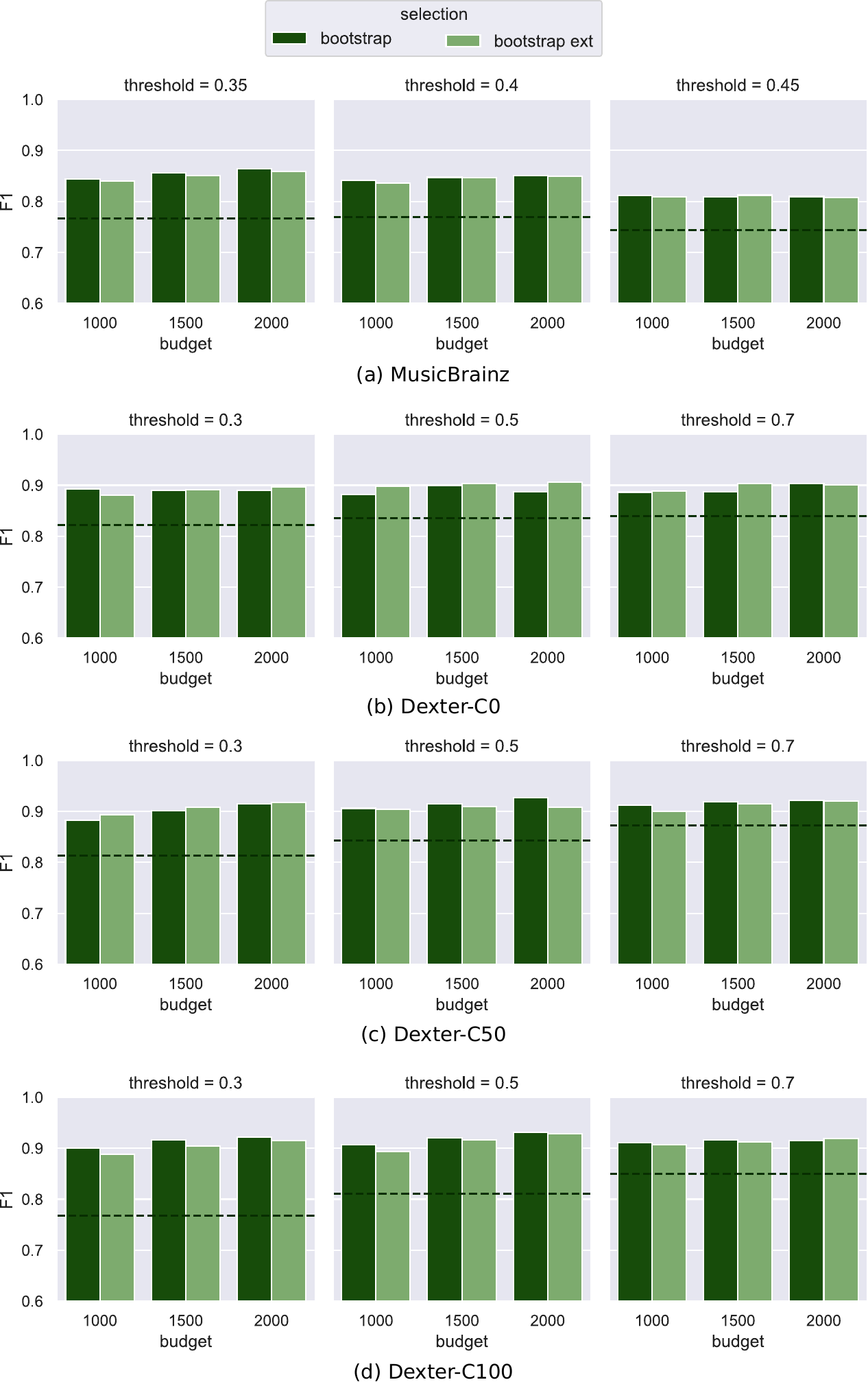}
    \caption{Results on Music Brainz and Dexter(C0, C50, C100) datasets with different duplicate ratios considering the basic selection strategy (bootstrap) and the cluster-specific selection (bootstrap ext) in the active learning step.}
    \label{fig:al_results}
\end{figure}

\subsection{Cluster-specific Training Data Selection}\label{subsec:al_mod}
Initially, we compare the cluster-specific active learning method (bootstrap ext) with the original approach~\cite{Moz14} (bootstrap) shown in~\Cref{fig:al_results}. Both selection strategies improve the initial quality considering the given similarity graph (dashed line) for all datasets. The results also show robustness using different labeling budgets since the F1-score only differs by $\approx$0.017 (bootstrap) and $\approx$0.016 (bootstrap ext). 

In terms of the effectiveness regarding data quality issues, the cluster-specific selection strategy slightly improves the results by up to $\approx$0.018 compared to the baseline for repairing clusters of dirty datasets such as C0. However, the baseline achieves better results for duplicate-free datasets than the extended selection. 

% \begin{figure}[!t]
%     \centering
%     \begin{subfigure}[t]{0.39\textwidth}
%         \includegraphics[width=\textwidth]{figures/comparison_C0.pdf}
%         \caption{Dexter-C0}
%     \end{subfigure}
%    \begin{subfigure}[t]{0.6\textwidth}
%         \includegraphics[width=\textwidth]{figures/comparison_C50.pdf}
%         \caption{Dexter-C50}
%     \end{subfigure}
%     \begin{subfigure}[b]{0.45\textwidth}
%         \includegraphics[width=\textwidth]{figures/comparison_C100.pdf}
%         \caption{Dexter-C100}
%     \end{subfigure}
%     \begin{subfigure}[b]{0.45\textwidth}
%         \includegraphics[width=\textwidth]{figures/comparison_MusicBrainz.pdf}
%         \caption{MusicBrainz}
%     \end{subfigure}
%     %\vspace{-0.115cm}
%     \caption{F1 score results of our proposed approach (GraphCR) as compared with the other repair methods CLIP~\cite{saaedi2018famer}, affinity propagation clustering (MSCD-AP)~\cite{LermSR2021} as well as agglomerative hierarchical clustering methods~\cite{SaeediDR2021} with the different variations regarding the merging step single (MSCD S-LINK), complete (MSCD C-LINK) and average (MSCD A-LINK).}
%     \label{fig:comparison}
% \end{figure}

\begin{figure}[tb]
    \centering
    \includegraphics[width=0.9\textwidth]{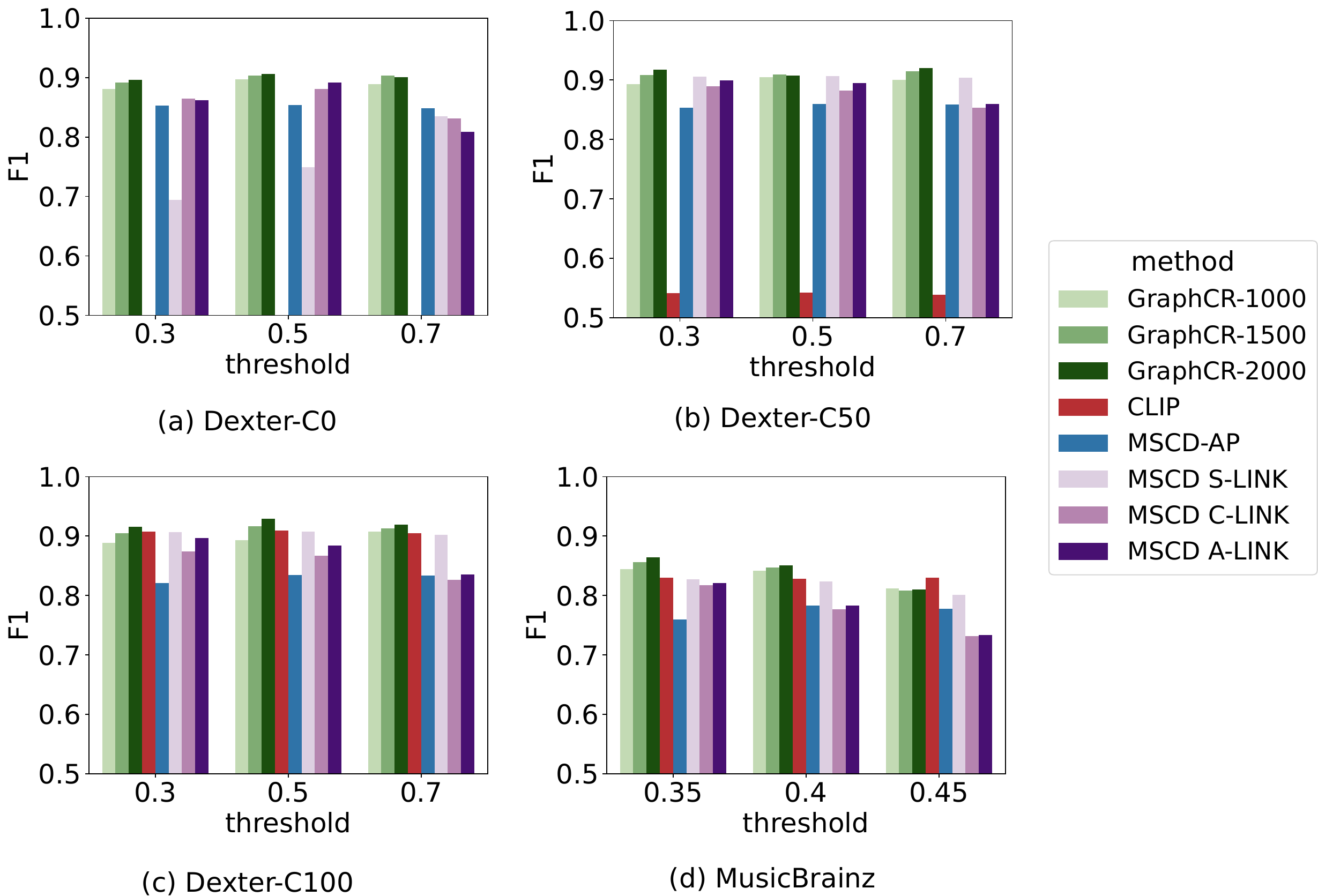}
    \caption{F1-score results of our proposed approach (GraphCR) as compared with the other repair methods CLIP~\cite{saaedi2018famer}, affinity propagation clustering (MSCD-AP)~\cite{LermSR2021} as well as agglomerative hierarchical clustering methods~\cite{SaeediDR2021} with the different variations regarding the merging step single (MSCD S-LINK), complete (MSCD C-LINK) and average (MSCD A-LINK).}
    \label{fig:comparison}
\end{figure}

\begin{figure}[tb]
     \centering
     \includegraphics[width=0.6\textwidth]{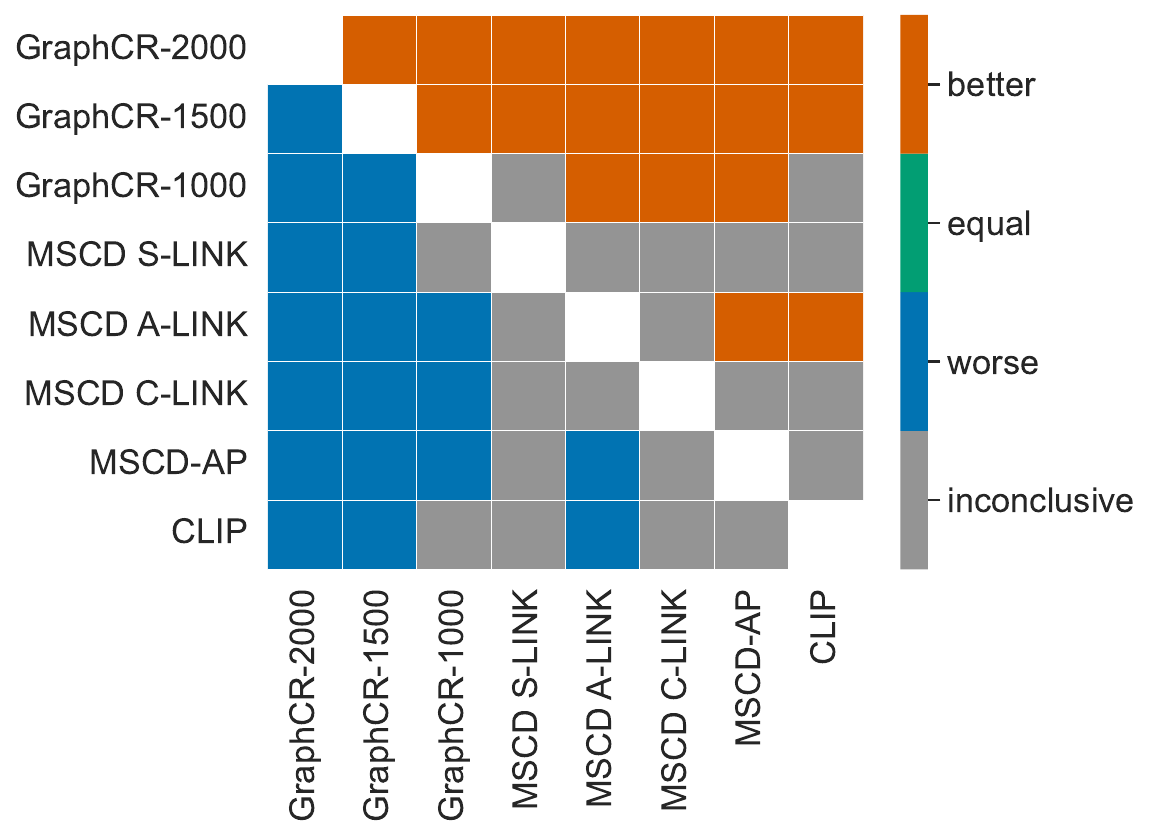}
     \caption{Decision matrix comparing cluster repair approaches using Bayesian signed rank tests. Each cell shows the decision when comparing the row approach to the column approach.}
     \label{fig:decision_matrix}
\end{figure}
\vspace{-0.02cm}

\subsection{Comparison with Existing Repair Methods}\label{subsec:comp}

We compare our method with the CLIP method~\cite{saaedi2018famer} and two clustering-based repair approaches based on hierarchical agglomerative clustering with the different clustering strategies \textit{single}(MSCD S-Link), \textit{complete}(MSCD C-Link), and \textit{average}(MSCD A-Link) \cite{SaeediDR2021} as well as affinity propagation(MSCD-AP)~\cite{LermSR2021}.

Depending on the datasets' dirtiness and the clustering algorithm's configuration, the existing methods improve the quality of the initial similarity graphs. However, the resulting quality of a certain configuration highly differs regarding the different datasets and thresholds. For instance, the F1-score achieved by CLIP ranges from 0.1(C0) to 0.9(C100). Even the results of the hierarchical clustering methods designed for mixed datasets differ between 0.83(C0) and 0.89(C100) for MSCD A-LINK. GraphCR outperforms the existing approaches concerning the used thresholds and datasets for labeling budgets $\geq$ 1500. The F1-score differences regarding the dirtiness of datasets are especially small compared to the baseline approaches. For instance, the F1-score difference is smaller than 0.03 for GraphCR concerning C0 and C100 using a labeling budget of 1500 overall thresholds. 

To properly compare the performance of these approaches, we rely on the Bayesian analysis proposed in~\cite{benavoli17time}.
To determine significant differences between the two approaches, we rely on a Bayesian signed rank test~\cite{benavoli14bayesian}.
Using Bayesian statistics gives the advantage over a frequentist hypothesis testing approach that can not only reject but also accept a null hypothesis.
We can also define a \textit{region of practical equivalence} (ROPE) where approaches perform equally well.
The \texttt{Autorank}~\cite{herbold20autorank} package provides these methods and can automatically set the ROPE in relation to effect size.
The result of our analysis then provides us with a probability that one approach is better (or worse) than another.
We provide a verdict if one of these probabilities is $\geq 95$\%, or else we see the result as inconclusive.

The result of this analysis is a decision matrix in Figure~\ref{fig:decision_matrix}.
We compared the F1-score of each approach across all dataset/threshold combinations. For GraphCR, we also compare different labeling budgets.
It is evident that GraphCR, with a labeling budget of 2000, is significantly better than all other approaches. Even with a budget of 1500, it is still better than all approaches that are not GraphCR.

\subsection{Effect of Noisy Similarities}\label{subsec:noise}

The LOD cloud consists of a large number of \texttt{sameAs} links. Nonetheless, a significant portion of these connections originates from sources whose correctness is suspect. This uncertainty can arise from various sources, including the utilization of improperly configured entity resolution systems or the presence of inaccurately defined links. Within this section, we aim to assess the robustness of our proposed method in the presence of noisy similarities. In the following experiment, we randomly select a specified ratio of edges and set the similarity to a random number between 0 and 1. We assume that the noisy similarity edges negatively impact the generated classification model and, therefore, the result of the cluster repair.  

% \begin{figure}[t]
%     \begin{subfigure}[b]{0.32\textwidth}
%      \includegraphics[width=\textwidth]{figures/MusicBrainz_0.35_bootstrap.pdf}
%      \caption{t=0.35}
%     \end{subfigure}
%     \begin{subfigure}[b]{0.32\textwidth}
%      \includegraphics[width=\textwidth]{figures/MusicBrainz_0.4_bootstrap.pdf}
%      \caption{t=0.4}
%     \end{subfigure}
%     \begin{subfigure}[b]{0.32\textwidth}
%     \includegraphics[width=\textwidth]{figures/MusicBrainz_0.45_bootstrap.pdf}
%      \caption{t=0.45}
%     \end{subfigure}
%     %\vspace{-0.115cm}
%     \caption{Results on Music Brainz with various error ratios of the similarities using the basic selection strategy in the active learning step.}
%     \label{fig:error_edges}
% \end{figure}

\begin{figure}[t]
    \begin{subfigure}[b]{\textwidth}
    \includegraphics[width=\textwidth]{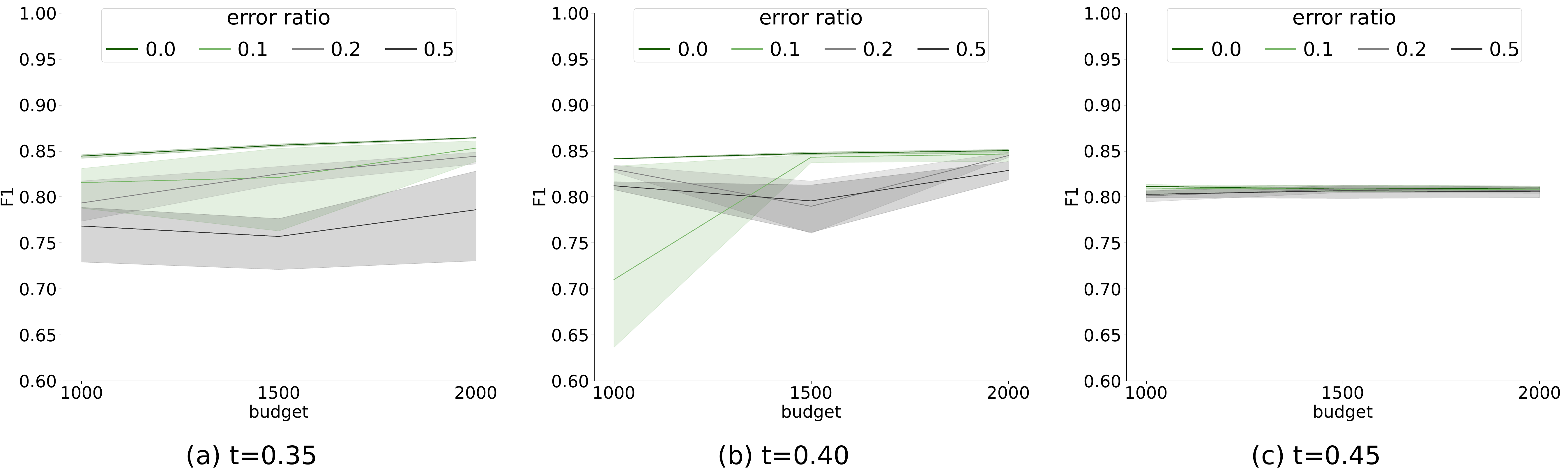}
    \caption{MusicBrainz\label{fig:error_edges_mb}}
    \end{subfigure}
    \begin{subfigure}[b]{\textwidth}
    \includegraphics[width=\textwidth]{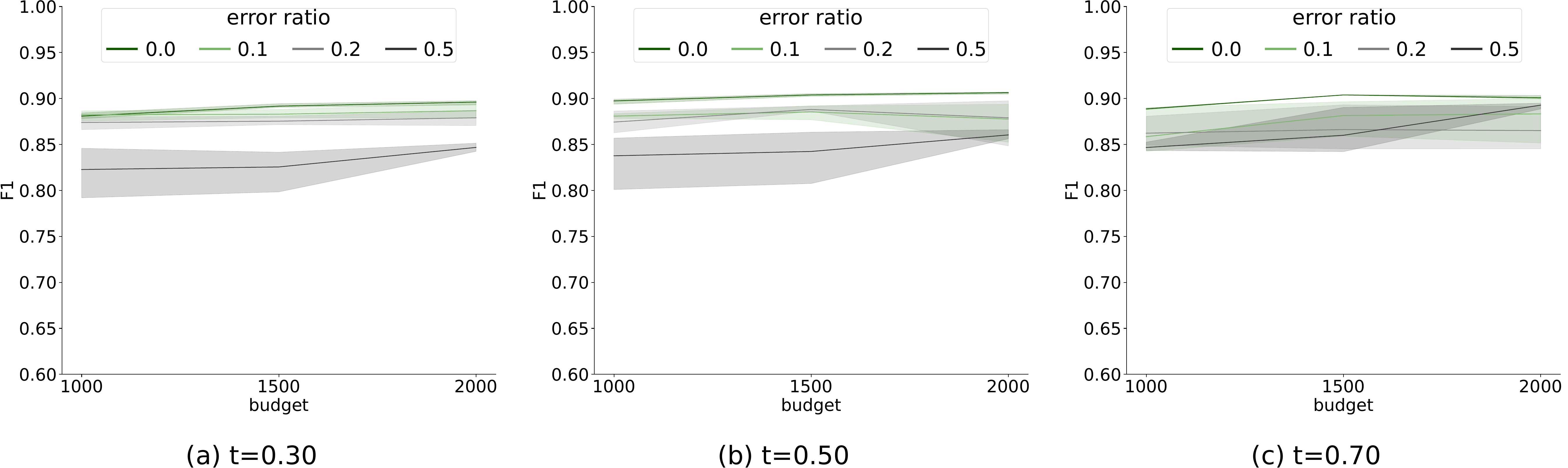}
    \caption{Dexter C0\label{fig:dexter_error_edges}}
    \end{subfigure}
    \caption{Results on Dexter and MusicBrainz datasets with various error ratios of the similarities using the extended resp. Basic selection strategy in the active learning step.}
    \label{fig:error_edges}
\end{figure}

\Cref{fig:error_edges} show the results for the MusicBrainz dataset applying the bootstrapping selection (bootstrap) and the Dexter dataset C0 using the extended strategy (bootstrap ext). Using the original similarity graph without any noise (error ratio = 0) for the MusicBrainz dataset results in a small quality difference of at most 0.007. As expected, the increase in the error ratio leads to a decrease in quality. Especially, an error ratio of 0.5 leads to F1 score differences up to 10\%(MusicBrainz) and 6\%(Dexter-C0). The threshold and the budget influence the robustness of the results. An increasing threshold and labeling budget positively influence the robustness. A higher threshold reduces the number of noisy similarities being already included in the original graph compared to a smaller threshold. Moreover, the increasing budget leads to a reduced impact of noisy links. In summary, the results showed that our graph-based cluster repair approach improves the quality of the initial clusters using a moderate labeling effort. Moreover, the method outperforms the existing cluster repair methods with less configuration effort than the existing approaches.

% TODO: A little conclusive sentence about the eval results$

\section{Conclusion}\label{sec:conclusion}
Cluster repair methods play an integral role in data integration tasks being relevant for knowledge graph completion. Knowledge graphs support comprehensive analysis and the realization of complex question-answering systems. The construction from various and heterogeneous data sources with different quality levels requires entity resolution identifying records representing the same entity. Due to the transitive closure of \texttt{sameAs} links and incorrect edges, the result potentially consists of clusters derived by the similarity graph with records representing different entities. Therefore, cluster repair methods aim to identify incorrect records in clusters utilizing the generated similarity graph. The majority of repair methods intensively rely on the assumption of duplicate-free data sources. Current methods try to overcome this issue by modifying general clustering approaches. However, the results vary depending on the configuration of the entity resolution tool and the degree of dirtiness.

Therefore, we proposed a novel cluster repair method relying on graph metric features enabling the training of a classification model. The computed model is used to modify the initial clusters by iteratively removing edges being classified as non-match. Due to the lack of available training data, we integrate and extend an active learning method by considering cluster-specific characteristics to guarantee that the selected training samples represent the complete dataset. We evaluated our proposed approach using two real-world datasets. The results showed that our method outperforms existing approaches using a moderate labeling budget. 

In future work, we plan to apply our method to datasets crawled from the LOD cloud and potentially improve the current state of link repositories. In terms of effectiveness, we plan to consider the relations of the underlying knowledge graphs and determine similarity edge features in combination with the semantic edges. Due to the manual labeling process in the active learning step, we will evaluate graph-based methods~\cite{Primpeli21graphAl} to reduce the number of selected samples. Moreover, we will also consider cluster-wise active learning strategies where a whole cluster is labeled regarding the correct and incorrect record. This allows the application of graph augmentation, such as the addition of correct links regarding one cluster.

\subsection*{Availability}
Reference code and datasets are available from our repository at \url{https://anonymous.4open.science/r/graphCR-DA60} and \url{https://www.dropbox.com/scl/fo/xtawtr2qbeckt0esu5v0h/h?rlkey=6lkz7hwsyhnqgyhg9quh7pp96&dl=0}.

\subsection*{Acknowledgments}
%This work was partially supported by lorem ipsum dolor sit amet, consetetur sadipscing elitr, sed diam nonumy eirmod tempor invidunt ut labore et dolore magna aliquyam erat, sed diam voluptua.
This work was partially supported by the German Federal Ministry of Education and Research by funding the "Center for Scalable Data Analytics and Artificial Intelligence Dresden/Leipzig" (ScaDS.AI).

\bibliographystyle{splncs04}
\bibliography{bibliography.bib}

\begin{thebibliography}{10}
\providecommand{\url}[1]{\texttt{#1}}
\providecommand{\urlprefix}{URL }
\providecommand{\doi}[1]{https://doi.org/#1}

\bibitem{Ara10}
Arasu, A., G{\"o}tz, M., Kaushik, R.: On active learning of record matching
  packages. In: ACM SIGMOD. pp. 783--794. Indianapolis (2010).
  \doi{10.1145/1807167.1807252}

\bibitem{Bel12}
Bellare, K., Iyengar, S., Parameswaran, A.G., Rastogi, V.: Active sampling for
  entity matching. In: ACM SIGKDD. pp. 1131--1139. Beijing (2012).
  \doi{10.1145/2339530.2339707}

\bibitem{benavoli17time}
Benavoli, A., Corani, G., Demsar, J., Zaffalon, M.: Time for a change: a
  tutorial for comparing multiple classifiers through bayesian analysis. J.
  Mach. Learn. Res.  \textbf{18},  77:1--77:36 (2017),
  \url{http://jmlr.org/papers/v18/16-305.html}

\bibitem{benavoli14bayesian}
Benavoli, A., Corani, G., Mangili, F., Zaffalon, M., Ruggeri, F.: A bayesian
  wilcoxon signed-rank test based on the dirichlet process. In: Proceedings of
  the 31th International Conference on Machine Learning, {ICML}. {JMLR}
  Workshop and Conference Proceedings, vol.~32, pp. 1026--1034. JMLR.org
  (2014), \url{http://proceedings.mlr.press/v32/benavoli14.html}

\bibitem{Chr12}
Christen, P.: Data Matching -- Concepts and Techniques for Record Linkage,
  Entity Resolution, and Duplicate Detection. Springer (2012).
  \doi{10.1007/978-3-642-31164-2}

\bibitem{christen19infoal}
Christen, V., Christen, P., Rahm, E.: Informativeness-based active learning for
  entity resolution. In: Machine Learning and Knowledge Discovery in Databases
  - International Workshops of {ECML} {PKDD}. Communications in Computer and
  Information Science, vol.~1168, pp. 125--141. Springer (2019).
  \doi{10.1007/978-3-030-43887-6\_11}

\bibitem{Doan20magellan}
Doan, A., Konda, P., C., P.S.G., Govind, Y., Paulsen, D., Chandrasekhar, K.,
  Martinkus, P., Christie, M.: Magellan: toward building ecosystems of entity
  matching solutions. Commun. {ACM}  \textbf{63}(8),  83--91 (2020).
  \doi{10.1145/3405476}, \url{https://doi.org/10.1145/3405476}

\bibitem{Harris08bridges}
Harris, J.M., Hirst, J.L., Mossinghoff, M.J.: Combinatorics and Graph Theory,
  Second Edition. Undergraduate Texts in Mathematics, Springer (2008)

\bibitem{herbold20autorank}
Herbold, S.: Autorank: {A} python package for automated ranking of classifiers.
  J. Open Source Softw.  \textbf{5}(48), ~2173 (2020).
  \doi{10.21105/JOSS.02173}, \url{https://doi.org/10.21105/joss.02173}

\bibitem{Hildebrandt2020}
Hildebrandt, K., Panse, F., Wilcke, N., Ritter, N.: Large-scale data pollution
  with apache spark. {IEEE} Trans. Big Data  \textbf{6}(2),  396--411 (2020).
  \doi{10.1109/TBDATA.2016.2637378}

\bibitem{Hofer2023ConstructionOK}
Hofer, M., Obraczka, D., Saeedi, A., Kopcke, H., Rahm, E.: Construction of
  knowledge graphs: State and challenges. arXiv preprint  (2023).
  \doi{https://doi.org/10.48550/arXiv.2302.11509}

\bibitem{LermSR2021}
Lerm, S., Saeedi, A., Rahm, E.: Extended affinity propagation clustering for
  multi-source entity resolution. In: Datenbanksysteme f{\"{u}}r Business,
  Technologie und Web ({BTW}). pp. 217--236 (2021). \doi{10.18420/btw2021-11}

\bibitem{Bing21Bert}
Li, B., Miao, Y., Wang, Y., Sun, Y., Wang, W.: Improving the efficiency and
  effectiveness for bert-based entity resolution. In: Thirty-Fifth {AAAI}
  Conference on Artificial Intelligence. pp. 13226--13233. {AAAI} Press (2021).
  \doi{10.1609/AAAI.V35I15.17562}

\bibitem{Moz14}
Mozafari, B., Sarkar, P., Franklin, M., Jordan, M., Madden, S.: Scaling up
  crowd-sourcing to very large datasets: A case for active learning. PVLDB
  Endowment  \textbf{8}(2),  125--136 (Oct 2014)

\bibitem{Mudgal2018deepLearning}
Mudgal, S., Li, H., Rekatsinas, T., Doan, A., Park, Y., Krishnan, G., Deep, R.,
  Arcaute, E., Raghavendra, V.: Deep learning for entity matching: {A} design
  space exploration. In: Das, G., Jermaine, C.M., Bernstein, P.A. (eds.)
  Proceedings of the 2018 International Conference on Management of Data. pp.
  19--34. {ACM} (2018). \doi{10.1145/3183713.3196926}

\bibitem{nentwig2017distributed}
Nentwig, M., Gro{\ss}, A., M{\"{o}}ller, M., Rahm, E.: Distributed holistic
  clustering on linked data. In: On the Move to Meaningful Internet Systems.
  {OTM} 2017 Conferences - Confederated International Conferences: CoopIS,
  C{\&}TC, and {ODBASE} 2017, Proceedings, Part {II}. Lecture Notes in Computer
  Science, vol. 10574, pp. 371--382. Springer (2017).
  \doi{10.1007/978-3-319-69459-7\_25}

\bibitem{NentwigHNR17survey}
Nentwig, M., Hartung, M., Ngomo, A.N., Rahm, E.: A survey of current link
  discovery frameworks. Semantic Web  \textbf{8}(3),  419--436 (2017).
  \doi{10.3233/SW-150210}, \url{https://doi.org/10.3233/SW-150210}

\bibitem{Newman2010NetworksAI}
Newman, M.E.J.: Networks: An introduction (2010),
  \url{https://api.semanticscholar.org/CorpusID:60557556}

\bibitem{Ngonga21limes}
Ngomo, A.N., Sherif, M.A., Georgala, K., Hassan, M.M., Dre{\ss}ler, K., Lyko,
  K., Obraczka, D., Soru, T.: {LIMES:} {A} framework for link discovery on the
  semantic web. K{\"{u}}nstliche Intell.  \textbf{35}(3),  413--423 (2021).
  \doi{10.1007/S13218-021-00713-X},
  \url{https://doi.org/10.1007/s13218-021-00713-x}

\bibitem{ngomo2014collibri}
Ngomo, A.N., Sherif, M.A., Lyko, K.: Unsupervised link discovery through
  knowledge base repair. In: The Semantic Web: Trends and Challenges - 11th
  International Conference, {ESWC} 2014, Proceedings. Lecture Notes in Computer
  Science, vol.~8465, pp. 380--394. Springer (2014).
  \doi{10.1007/978-3-319-07443-6\_26}

\bibitem{Ngo12}
Ngonga~Ngomo, A.C., Lyko, K.: Eagle: Efficient active learning of link
  specifications using genetic programming. In: The Semantic Web: Research and
  Applications. pp. 149--163. Berlin, Heidelberg (2012)

\bibitem{Pan23KGLLM}
Pan, S., Luo, L., Wang, Y., Chen, C., Wang, J., Wu, X.: Unifying large language
  models and knowledge graphs: {A} roadmap. arXiv preprint  (2023).
  \doi{10.48550/ARXIV.2306.08302}

\bibitem{Peters2023GPT}
Peeters, R., Bizer, C.: Using {ChatGPT} for entity matching. In: Abell{\'{o}},
  A., Vassiliadis, P., Romero, O., Wrembel, R., Bugiotti, F., Gamper, J.,
  Vargas{-}Solar, G., Zumpano, E. (eds.) New Trends in Database and Information
  Systems - {ADBIS} 2023. Communications in Computer and Information Science,
  vol.~1850, pp. 221--230. Springer (2023). \doi{10.1007/978-3-031-42941-5\_20}

\bibitem{Primpeli21graphAl}
Primpeli, A., Bizer, C.: Graph-boosted active learning for multi-source entity
  resolution. In: The Semantic Web - {ISWC} 2021 - 20th International Semantic
  Web Conference, {ISWC} 2021, Virtual Event, October 24-28, 2021, Proceedings.
  Lecture Notes in Computer Science, vol. 12922, pp. 182--199. Springer (2021).
  \doi{10.1007/978-3-030-88361-4\_11}

\bibitem{SaeediDR2021}
Saeedi, A., David, L., Rahm, E.: Matching entities from multiple sources with
  hierarchical agglomerative clustering. In: {IC3K}. pp. 40--50. {SCITEPRESS}
  (2021). \doi{10.5220/0010649600003064}

\bibitem{saaedi2018famer}
Saeedi, A., Peukert, E., Rahm, E.: Using link features for entity clustering in
  knowledge graphs. In: The Semantic Web - 15th International Conference,
  {ESWC} 2018, Proceedings. Lecture Notes in Computer Science, vol. 10843, pp.
  576--592. Springer (2018). \doi{10.1007/978-3-319-93417-4\_37}

\bibitem{SaeediER2020}
Saeedi, A., Peukert, E., Rahm, E.: Incremental multi-source entity resolution
  for knowledge graph completion. In: {ESWC}. vol. 12123, pp. 393--408.
  Springer (2020). \doi{10.1007/978-3-030-49461-2_23}

\bibitem{shen08Soccer}
Shen, W., DeRose, P., Vu, L.H., Doan, A., Ramakrishnan, R.: Source-aware entity
  matching: {A} compositional approach. In: Chirkova, R., Dogac, A.,
  {\"{O}}zsu, M.T., Sellis, T.K. (eds.) Proceedings of the 23rd International
  Conference on Data Engineering, {ICDE} 2007, The Marmara Hotel, Istanbul,
  Turkey, April 15-20, 2007. pp. 196--205. {IEEE} Computer Society (2007).
  \doi{10.1109/ICDE.2007.367865}

\bibitem{Yang23GPTKG}
Yang, L., Chen, H., Li, Z., Ding, X., Wu, X.: {ChatGPT} is not enough:
  Enhancing large language models with knowledge graphs for fact-aware language
  modeling. arXiv preprint  (2023). \doi{10.48550/ARXIV.2306.11489}

\end{thebibliography}
\end{document}